\documentclass[11pt, a4paper, logo, onecolumn,copyright,most]{tmlrmeljjt}

\usepackage[authoryear, sort&compress, round]{natbib}
\makeatletter
\providecommand{\captionlabel}{\@ifnextchar\bgroup\captionlabel@i\captionlabel@ii}
\newcommand{\captionlabel@i}[1]{}
\newcommand{\captionlabel@ii}{}
\makeatother

\title{Does Execution Require Target KV Fidelity? A Mixed-Fidelity KV Runtime for LLM Serving}

\usepackage{enumitem}
\usepackage{wrapfig}
\usepackage{graphicx}
\usepackage{subcaption}
\usepackage{booktabs}
\usepackage[ruled,vlined]{algorithm2e}
\usepackage{multirow}
\usepackage{xcolor}
\usepackage{hyperref}
\hypersetup{
    colorlinks=true,
    citecolor=teal,
    linkcolor=red,
}

\author[$\diamondsuit$]{Jiantong Jiang}
\author[$\heartsuit$]{Yue Yang}
\author[$\diamondsuit$]{Peiyu Yang}
\author[$\diamondsuit$]{Feng Liu}

\affil[ \hspace{-0.2em}]{$\diamondsuit$ University of Melbourne
}
\affil[ \hspace{-0.2em}]{$\heartsuit$ Maincode}

\correspondingauthors{fengliu.ml@gmail.com}

\begin{document}

\begin{abstract}
Large language model (LLM) serving is increasingly constrained by the GPU memory consumed by key-value (KV) caches. Existing compression, eviction, and offloading techniques alleviate this pressure, but serving runtimes typically treat only the configured target KV representation as execution-ready. Under memory pressure, this target-only contract can turn KV shortage into request stalls and preemptions. We present ElasticKV, a mixed-fidelity KV runtime built on the observation that target fidelity need not gate execution. ElasticKV introduces a compact intermediate KV state, making fidelity a runtime-managed execution property. To realize this state in a paged serving runtime, ElasticKV combines (i) a pair-structured layout that turns fidelity reduction into reusable GPU capacity, (ii) a dual-mode attention backend that directly consumes the compact state while preserving the native target-only path, and (iii) pressure-aware fidelity management that adapts KV fidelity to memory pressure. Our extensive evaluation across diverse workloads, model families and scales, and GPU platforms demonstrates the effectiveness and generality of ElasticKV. Under high concurrency, ElasticKV achieves 3.8-4.0$\times$ lower time-to-first-token (TTFT) and 9.1$\times$ lower P90 TTFT than vLLM while preserving generation quality. 
\end{abstract}

\maketitle

\section{Introduction}

As large language model (LLM) services scale to longer contexts and higher request concurrency, \textit{key-value (KV) cache} increasingly determines how much workload a GPU can sustain~\citep{miao2025towards, jiang2026towards}. During autoregressive generation, each active request continuously accumulates KV states, growing its memory footprint~\citep{zhao2026survey}. As this demand approaches GPU capacity, even a modest increase can push the serving system into a different regime, triggering request waiting, preemption, or context recomputation and causing sharp latency degradation.



%
Existing work alleviates KV memory pressure by compressing KV cache, sparsifying or evicting less important KV entries, or moving KV cache across memory tiers~\citep{kivi, commvq, h2o, attentionsink, lserve, flexgen, infinigen, lmcache}. These techniques improve how much KV can be stored, reused, or transferred. 

\begin{figure}[t]
\centerline{\includegraphics[width=\textwidth]{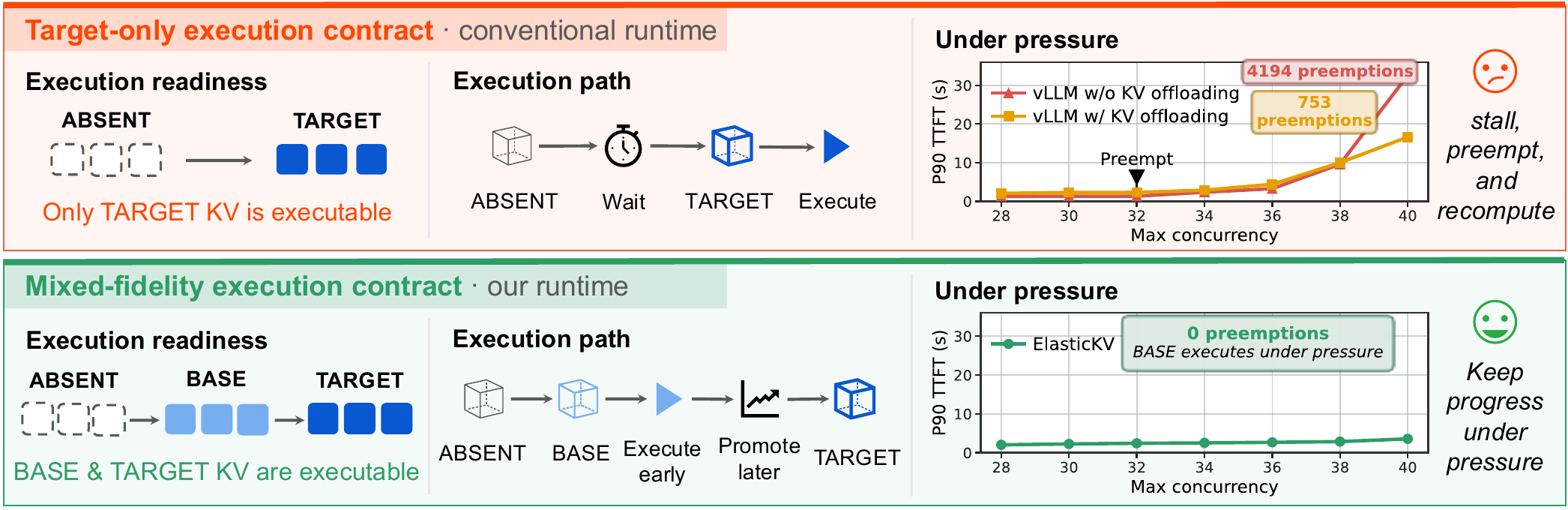}}
	\centering
    \vspace{-1mm}
	\caption{Comparison of target-only execution contract and mixed-fidelity execution contract. }
	\label{fig_motivation}
    \vspace{-3mm}
\end{figure}

However, these works overlook a different runtime constraint: \textit{when is a KV state sufficient for request execution?} In conventional serving, request readiness is tied to the runtime's configured \textit{target fidelity}. 
%
Figure~\ref{fig_motivation} makes this coupling concrete. From an execution-readiness view, the required KV is either target-ready on GPU (i.e., \textit{TARGET}) or not executable (i.e., \textit{ABSENT}). If TARGET is not available under pressure, the runtime may wait or preempt requests, potentially requiring later recomputation. As concurrency rises, this boundary can turn a KV shortage into a preemption-driven P90 time-to-first-token (TTFT) knee. This suggests that the bottleneck is not only how much KV can be stored or moved, but also which KV representations the runtime can accept for execution.
%

Our key observation is that \textit{reaching target KV fidelity need not gate execution}. We therefore introduce \textit{BASE}, a compact intermediate GPU representation that is directly executable and can later be refined exactly to TARGET. Under this mixed-fidelity execution contract, BASE lets requests execute with a smaller KV footprint when resources are constrained. Since BASE trades fidelity for capacity and may incur a quality cost, TARGET remains the preferred state and is restored when resources permit. As shown in Fig.~\ref{fig_motivation}, BASE execution avoids the preemption-driven knee in the representative pressure case and keeps P90 TTFT substantially more stable as concurrency rises. 




While conceptually simple, this shift is substantially harder to realize in a paged serving runtime due to three key challenges.
(i) \textbf{Internal fragmentation.} Paged KV managers allocate memory in fixed-size physical blocks~\citep{vllm}. Therefore, lowering KV fidelity within a physical block does not by itself release reusable GPU capacity. (ii) \textbf{Execution mismatch.} An intermediate representation offers no progress benefit if request readiness and attention execution still require TARGET~\citep{lynx, cachegen}, while mixed-fidelity support should not penalize the common TARGET-only path. (iii) \textbf{Transition asymmetry.} TARGET-to-BASE transitions release GPU capacity and must reclaim enough to avoid disruption, whereas BASE-to-TARGET transitions consume capacity and should occur only when sufficient headroom is available.
Together, these challenges require coordinated runtime support across memory management, execution, and online control, rather than a standalone KV compression choice.


ElasticKV addresses these challenges with three coordinated mechanisms: (i) a pair-structured KV layout to make compact KV capacity \textit{physically reclaimable}, (ii) a dual-mode elastic attention backend to make BASE \textit{directly executable} while preserving the native TARGET path, and (iii) pressure-aware fidelity control that \textit{dynamically manages} asymmetric demotion and promotion under memory pressure.
%
We implement ElasticKV on top of vLLM~\citep{vllm} and evaluate it across diverse workloads, models, and GPU platforms. ElasticKV offers its largest gains near capacity boundaries, where limited BASE exposure can avoid more expensive request preemptions. Under high concurrency, ElasticKV eliminates request preemptions and achieves 3.8-4.0$\times$ lower mean TTFT and 9.1$\times$ lower P90 TTFT than standard vLLM, while largely preserving generation quality. The benefits persist on realistic workloads, model scales, and both NVIDIA and AMD GPU platforms.
In summary, this paper makes the following main contributions. 
\begin{itemize}[leftmargin=*,itemsep=2pt,topsep=3pt,parsep=0pt]
    \item We identify target-fidelity-gated readiness as an overlooked runtime constraint under memory pressure and introduce a mixed-fidelity execution contract that decouples execution from target-fidelity residency.

    \item We design ElasticKV, a paged mixed-fidelity runtime that makes the compact intermediate KV states physically reclaimable, directly executable, and dynamically managed without slowing down the native TARGET path.

    \item We implement ElasticKV and show that it significantly reduces latency under pressure and expands the runnable serving range while largely preserving generation quality.
\end{itemize}

\section{Background and Related Work}




\subsection{Paged KV Cache Management}
\label{sec_paged}

During autoregressive generation, each active request maintains a \textit{key-value (KV) cache} of previously computed attention states, whose memory footprint grows as the sequence progresses. We provide additional background on LLM inference and KV cache lifecycle in App.~\ref{app_bg}.
Modern serving engines commonly manage KV memory through paged KV allocation. In vLLM’s PagedAttention~\citep{vllm}, KV is partitioned into fixed-size physical blocks, while each request maintains logical blocks mapped to physical GPU blocks through a block table. Attention follows this mapping to access non-contiguous physical blocks while preserving logical token order.

This paged abstraction has since become a common serving substrate and has been extended to KV offloading~\citep{lmcache, vllmoffload}, scheduling~\citep{fastswitch, gimbal}, block-structured attention backends~\citep{flashinfer}, and block-level KV reduction~\citep{pagedeviction, zipage, thinkv, kvcompress, lserve, diffkv}. ElasticKV retains this substrate, but revisits the execution contract that determines when a KV state is sufficient for request progress.

\subsection{KV Cache Optimization and Execution Contracts}
\label{sec_motivation}

\textbf{KV cache offloading and reuse.} Existing systems expand effective serving capacity by migrating KV cache across GPU and lower memory tiers~\citep{flexgen, infinigen, cachedattention, lmcache, cachegen, fastswitch, vllmoffload}. They primarily improve where KV resides and how efficiently it is transferred, while execution typically resumes after the required target representation is available on GPU. \citet{lynx} relax this requirement to overlap KV transfer with execution, establishing that KV availability need not be static. ElasticKV instead focuses on dynamically managing the fidelity of live GPU KV under resource pressure.

\textbf{KV cache reduction and system support.} A broad line of work aims to reduce the size or amount of retained KV. Quantization reduces KV precision~\citep{kivi, kvquant, zipcache, gear, ott, rotatekv, cq, vqllm, nsnquant, kvtuner, moqae, kvmix, pmkvq}, while sparsification and eviction retain or attend to selected KV entries~\citep{h2o, attentionsink, adakv, diffkv, pagedeviction, thinkv, snapkv, pyramidkv, lacache}. Recent systems further co-design KV reduction with runtime or kernel support~\citep{qserve, lserve}, showing that memory savings must be exposed to the runtime to yield end-to-end gains. ElasticKV shares this systems view, but does not propose a new compression or selection algorithm. Instead, it uses fidelity transition as a runtime control action.

\textbf{Execution contracts.} ElasticKV focuses on when a KV state is sufficient for forward progress. Under a \textit{target-only execution contract}, the configured TARGET can itself be quantized or mixed-precision, but remains the only execution-ready state. The key distinction is therefore not precision level, but whether target fidelity gates execution. ElasticKV instead introduces a \textit{mixed-fidelity execution contract} by making both BASE and TARGET executable and coordinating fidelity changes across physical capacity, readiness, and attention execution. Thus, ElasticKV complements existing KV migration and reduction techniques by making fidelity itself a runtime-managed execution state.

\section{ElasticKV Design and Implementation}

\begin{wrapfigure}{r}{.54\textwidth}
\centerline{\includegraphics[width=1.0\linewidth]{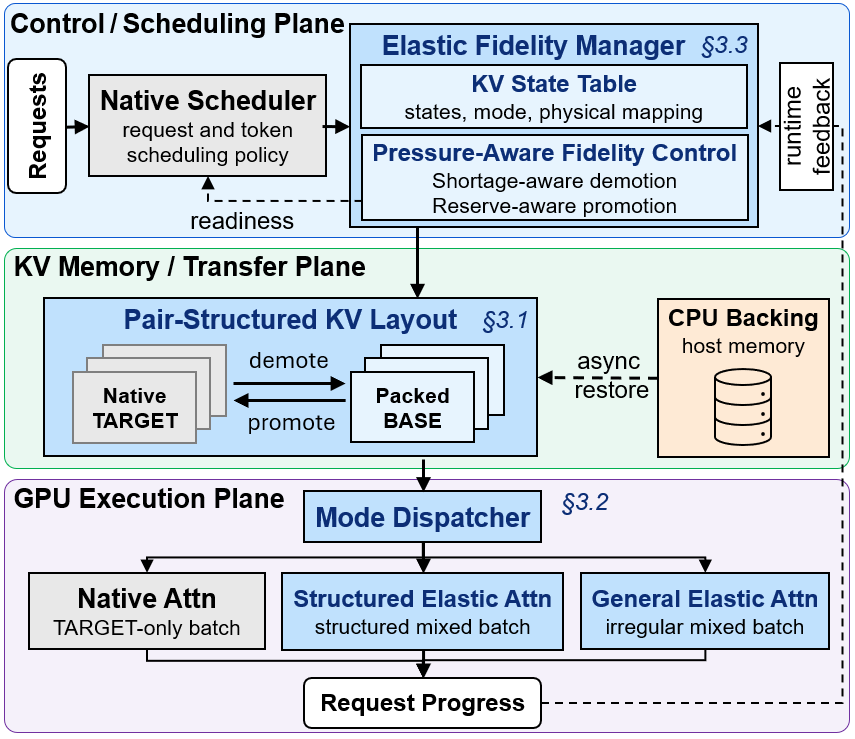}}
	\centering
    \vspace{-1mm}
	\caption{Overview of ElasticKV runtime architecture across control, KV memory, and GPU execution planes.} 
    \label{fig_architecture}
\vspace{-4mm}
\end{wrapfigure}

%
ElasticKV realizes the mixed-fidelity execution contract by extending the existing paged KV serving substrate with fidelity-aware memory management and execution support. 
Figure~\ref{fig_architecture} shows the overall architecture. In the control plane, ElasticKV retains vLLM's native scheduling policy, while an elastic fidelity manager
extends KV readiness and capacity management by tracking KV states and driving fidelity transitions (cf. \S\ref{sec_demo_promo}). These transitions are materialized through the pair-structured KV layout in the KV memory plane (cf. \S\ref{sec_layout}). The resulting KV state metadata then guides dispatch between native and elastic paths in the GPU execution plane, with mixed-fidelity batches further dispatched based on mode regularity 
(cf. \S\ref{sec_attn}).
Request and transition outcomes are fed back to inform scheduling decisions in subsequent steps.


In our implementation, TARGET is the native 16-bit KV representation, while BASE retains its high-order 8 bits. The remaining low-order 8 bits refine BASE exactly back to the original TARGET representation, without quantization parameters or a separate dequantization transform. BASE is therefore not an independently compressed representation, but an intermediate runtime state whose fidelity is reflected in allocator-visible capacity, request readiness, and attention execution. In what follows, we describe how ElasticKV makes mixed-fidelity KV physically reclaimable (cf. \S\ref{sec_layout}), directly executable (cf. \S\ref{sec_attn}), and dynamically managed under memory pressure (cf. \S\ref{sec_demo_promo}).


\subsection{Pair-Structured Elastic KV Layout}
\label{sec_layout}

\begin{wrapfigure}{r}{.32\textwidth}
\vspace{-4.5mm}
\centering
\centering
	\subfloat[Native block-based layout]
	{\includegraphics[width=\linewidth]{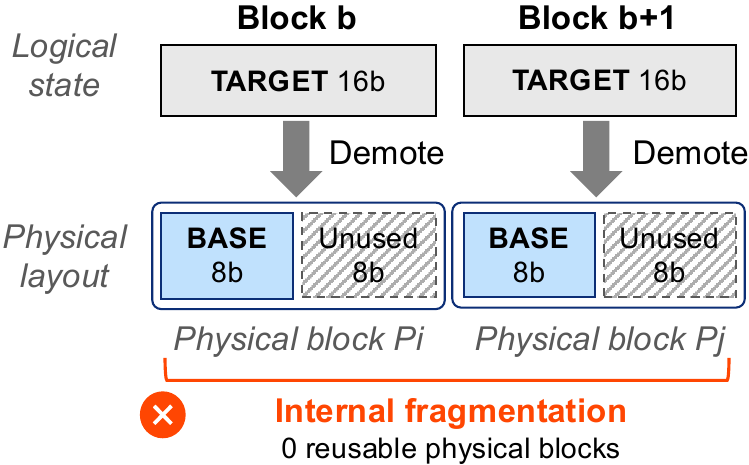}\label{fig_layout1}}
	
    \subfloat[Elastic pair-based layout]
	{\includegraphics[width=\linewidth]{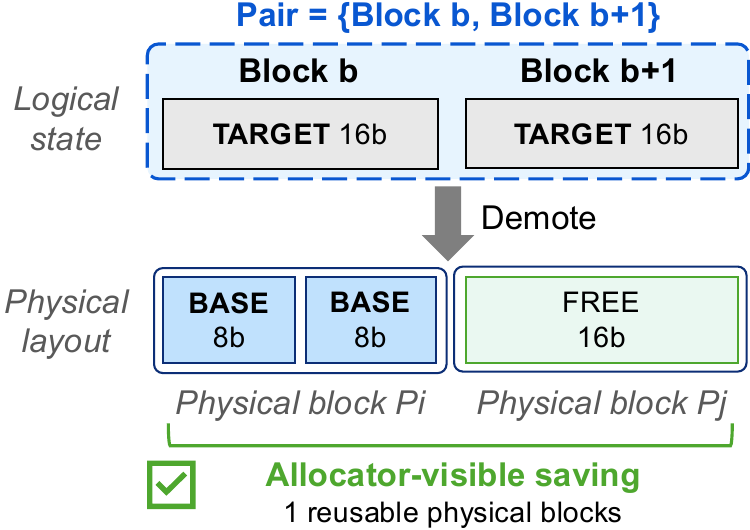}\label{fig_layout2}}
    \vspace{-1mm}
	\caption{Pair-structured KV layout for reclaimable BASE capacity.}  
	\label{fig_layout} 
\vspace{-4.5mm}
\end{wrapfigure}
BASE can relieve real memory pressure only if its compact representation translates into reusable GPU capacity. In ElasticKV, TARGET uses a 16-bit representation and BASE uses 8 bits. Ideally, this should translate into a reduction in physical occupancy. However, the paged KV substrate allocates memory in fixed-size physical blocks (cf. \S\ref{sec_paged}). If each logical block remains coupled to one physical block, storing the half-sized BASE state leaves the physical allocation unchanged, as illustrated in Fig.~\ref{fig_layout1}, causing internal fragmentation and stranding the unused half of the block. 


Mixed-fidelity KV therefore requires finer-grained physical management. However, simply partitioning KV blocks into arbitrarily small segments is also undesirable. Although finer segmentation increases flexibility, it can sacrifice the page-level regularity that makes native KV allocation, transfers, metadata construction, and attention execution efficient. ElasticKV must balance two requirements. First, the capacity saved by a compact representation should be returned as whole physical blocks. Second, the layout should preserve the page-level regularity needed by the runtime.


ElasticKV addresses this trade-off by managing mixed-fidelity storage at a fixed container granularity. Because BASE occupies exactly half the TARGET footprint, two logical blocks form the smallest regular container whose BASE states fit in one target-sized physical block. As shown in Fig.~\ref{fig_layout2}, a TARGET pair occupies two physical blocks. After demotion, the two BASE representations are packed into one block and the other is returned to the GPU KV pool. In this way, ElasticKV converts two half-block savings into one allocator-visible free block.
This pair granularity sits in a middle ground between fixed target-sized block allocation and fully segmented storage and thus satisfies both design requirements: it converts fidelity reduction into real capacity while preserving the page-level regularity of the native substrate.

Importantly, pairing affects only physical KV organization. Each pair member remains a standard logical block exposed to the serving runtime. This provides \textbf{native path compatibility}, since TARGET pairs preserve the native paged layout and require no layout conversion on the native execution path. Pairing also provides \textbf{state transition support}, making transitions bounded and reversible. TARGET-to-BASE demotion releases exactly one physical block per pair, while BASE-to-TARGET promotion reacquires one block and materializes the pair back.



\subsection{Dual-mode Elastic Attention Backend}
\label{sec_attn}

The pair-structured layout makes BASE physically reclaimable, but this capacity benefits request progress only if attention can consume BASE directly. At the same time, BASE execution should avoid penalizing the common TARGET-only case. ElasticKV therefore uses two top-level execution modes. TARGET-only batches follow the native attention path, while batches containing BASE enter the elastic path. A batch-level indicator lets TARGET-only batches bypass pair-mode processing, while mixed batches carry the pair-mode metadata needed by elastic execution. This preserves native execution when elasticity is unnecessary, while enabling mixed-fidelity execution under pressure.

Within the elastic attention path, ElasticKV represents each sequence's attention-visible KV as an ordered sequence of pair modes, marked as TARGET or BASE. ElasticKV exploits the regularity of these mode sequences before kernel execution. Its demotion protection rule keeps attention-sink~\citep{attentionsink} and recent-token pairs at TARGET, while eligible middle pairs may transition to BASE, so mixed sequences commonly form a TARGET-BASE-TARGET layout with at most two mode boundaries. In this common case, per-sequence boundary metadata compactly encodes the layout, allowing the structured elastic path to avoid dense per-pair mode lookup. Irregular sequences instead use an explicit pair-mode table and fall back to the general elastic path.

After dispatch, TARGET and BASE pairs use different load semantics within a common attention interface. TARGET pairs retain the native layout and load path. BASE pairs are loaded directly from the packed high-byte representation. The stored high-order bits are placed in their native positions, while the missing low-order bit positions are zero-filled to form attention values in the native datatype. Thus, BASE is consumed without materializing the missing refinement bits. Both paths produce the same output interface, leaving upper model layers agnostic to the underlying KV fidelity. Newly appended KV remains in the protected recent region and is therefore written in TARGET form, becoming eligible for demotion only later.

\subsection{Pressure-Aware Fidelity Management}
\label{sec_demo_promo}

Once BASE is physically reclaimable and executable, ElasticKV manages the asymmetric fidelity transitions: demotion releases GPU capacity and protects request progress under shortage, while promotion consumes capacity and restores the fidelity. ElasticKV thus makes demotion shortage-driven and promotion reserve-aware, as summarized in Alg.~\ref{alg_demote}.

\textbf{Shortage-aware proactive demotion.}
ElasticKV performs demotion before allocation failure. A side-effect-free dry run of the native scheduling logic probes the block demand $D_t$ of the current step, yielding a shortage $\Delta_t=\max(0,D_t-F_t)$ for the current free capacity $F_t$. If $\Delta_t=0$, the native allocation path is unchanged. Otherwise, ElasticKV selects eligible TARGET pairs for demotion. It excludes protected attention-sink and recent-token pairs and pairs without recoverable CPU backing, and prioritizes the remaining pairs using lightweight runtime signals. Each demoted pair releases one physical block via the pair layout in \S\ref{sec_layout} and remains executable in BASE.

\begin{wrapfigure}{r}{0.48\textwidth}
\vspace{-4mm}
\begin{algorithm}[H]
\caption{Pressure-aware fidelity control}
\label{alg_demote}
\KwIn{scheduler state $S_t$, pair metadata $M_t$, number of free blocks $F_t$, reserve $R_{t-1}$}
$D_t \gets \text{probe\_demand}(S_t)$\\
$\Delta_t \gets \max(0,D_t-F_t)$\\
$E_t \gets 0$\\
\If{$\Delta_t > 0$}{
    $\mathcal{P}_d \gets
    \text{eligible\_target\_pairs}(M_t,S_t)$\\
    $n_d \gets
    \min\!\left(
        |\mathcal{P}_d|,
        \max(\Delta_t,B_{\min}^{d})
    \right)$\\
    $\mathcal{B}_d \gets
    \text{rank\_and\_batch}(\mathcal{P}_d,n_d)$\\
    $\text{commit\_demotion}(\mathcal{B}_d)$\\
    $E_t \gets
    \max(0,|\mathcal{B}_d|-\Delta_t)$
}
\tcp{After required scheduling}
$R_t \gets
\max(0,R_{t-1}-\lambda)+E_t$\\
$P_t \gets
\max(0,F_t^{\mathrm{post}}-R_t)$\\
\If{$P_t>0$ \textbf{and} transfer slack is available}{
    $\mathcal{B}_p \gets
    \text{eligible\_base\_pairs}(M_t,P_t)$\\
    $\text{async\_promote}(\mathcal{B}_p)$
}
\end{algorithm}
\vspace{-3mm}
\end{wrapfigure}
In practice, however, demoting exactly $\Delta_t$ pairs can create many tiny transitions when shortages are small but frequent. ElasticKV therefore demotes at least $B_{\min}^{d}$ pairs once triggered. Any capacity released beyond the current shortage is recorded as surplus $E_t$ and remains in the free pool to amortize demotion over subsequent steps. 


\textbf{Reserve-aware asynchronous promotion.}
Promotion serves the complementary role of refining BASE back to TARGET, but consumes one additional GPU block per pair. ElasticKV treats promotion as an opportunistic fidelity recovery rather than progress-critical work. It is considered only after required scheduling and allocation and only when transfer slack is available.

However, eagerly promoting whenever a free block appears can immediately consume the GPU capacity recovered by recent demotion and induce promote-demote oscillation. To preserve recently created headroom from batched demotion, ElasticKV keeps a decaying demotion reserve $R_t$:
\[
    R_t = \max(0, R_{t-1}-\lambda) + E_t,
    \qquad
    P_t = \max(0, F_t^{\mathrm{post}} - R_t),
\]
where $\lambda$ is the per-step decay, $F_t^{\mathrm{post}}$ is the free capacity after required scheduling, and $P_t$ is the promotion budget. The reserve $R_t$ protects recently created surplus capacity and gradually returns to promotion as it decays. Eligible pairs are promoted asynchronously from CPU backing. Promotion is non-blocking. BASE remains execution-ready until TARGET has been fully materialized. 



\subsection{Implementation}
\label{sec_impl}

We implement ElasticKV on top of vLLM v0.15.0~\citep{vllm}, with runtime control in Python and attention kernels in Triton~\citep{triton}. This section provides an overview of implementation and lower-level details are provided in App.~\ref{app_impl}.  ElasticKV adds three integration points to vLLM across KV state management, transitions, and attention execution. 

\textbf{State and scheduler integration.} We augment \texttt{KVCacheManager} with a \texttt{LogicalBlockStateTable} that tracks fidelity state, physical mappings, CPU backing, and transition status while preserving vLLM's logical block namespace and native scheduling policy. A pre-allocation hook performs the side-effect-free demand probe in \S\ref{sec_demo_promo}, and BASE-ready requests remain executable without waiting for TARGET refinement. 

\textbf{Transition protocols.} We implement demotion as a synchronous in-place transition. Selected TARGET pairs are packed on GPU, and the freed page is returned to the block pool only after GPU completion. Our implementation refines BASE to TARGET asynchronously, with BASE remaining executable while refinement is in progress. The new mapping is published only after the target representation has been fully materialized, avoiding partially updated states. For tensor-parallel execution, pair transitions are coordinated across ranks so that they commit a consistent pair state.

\textbf{Elastic attention backend.} The registered \texttt{ELASTIC\_TRITON\_ATTN} backend receives sparse per-request BASE metadata that workers lower into block tables and execution metadata. TARGET-only batches invoke the native Triton attention path, while batches containing BASE blocks use the specialized elastic path described in \S\ref{sec_attn}. BASE values are loaded from the packed high-byte representation and expanded in registers through shift-and-bitcast operations before attention computation. 

\section{Evaluation}

\subsection{Evaluation Setup}
\label{sec_setup}

\noindent\textbf{Models and environments.} We evaluate ElasticKV on Llama-3.1-8B-Instruct~\citep{llama31}, Qwen3-8B~\citep{qwen3}, and Llama-3.1-70B-Instruct. Our main experiments run on an NVIDIA A100 GPU with 80GB HBM. To evaluate multi-GPU scalability and cross-vendor portability, we further run experiments on a server with 8 AMD MI355X GPUs, each with 288GB HBM.

\noindent\textbf{Datasets and workloads.} 
We evaluate generation quality on LongBench~\citep{longbench}. For controlled serving evaluation, we use 600 synthetic requests with 4096 input tokens and 1024 output tokens. We further use ShareGPT~\citep{sharegpt} and production traces from Mooncake~\citep{mooncake} to evaluate heterogeneous request lengths and time-varying serving pressure.

\noindent\textbf{Baselines.} We compare ElasticKV against two exact-fidelity serving regimes. Specifically, (i) \textit{Full-GPU} keeps FP16 KV entirely on GPU and preempts requests when GPU KV capacity is exhausted. (ii) \textit{Exact-offload} extends the effective capacity by moving KV between GPU and host memory, but requires an offloaded block to return to GPU before the request can resume. We instantiate the baselines using vLLM~\citep{vllm} and its KV offloading support~\citep{vllmoffload}. Exact-offload and ElasticKV use the same host-memory budget within each comparison.

\noindent\textbf{Metrics.} We measure generation quality using the official LongBench metrics~\citep{longbench}. For serving performance, we report \textit{time-to-first-token} (TTFT), \textit{time-per-output-token} (TPOT), \textit{throughput}, request preemptions, together with \textit{tail latency} P90 TTFT and P90 TPOT. For the Mooncake trace replay, we further follow its SLO-oriented \textit{effective request ratio}~\citep{mooncake} and characterize temporal behavior using queueing, drain, and KV-residency statistics.

\subsection{Evaluation on Generation Quality}
\label{sec_quality}

\begin{wraptable}{r}{0.38\textwidth}
\vspace{-5mm}
    \small
    \centering
    \caption{Average LongBench scores over 21 tasks. ElasticKV uses BASE-exposure caps. All-BASE is a 100\% BASE stress endpoint. Higher is better.}
    \vspace{-1mm}
    \label{tab_quality}
    \setlength{\tabcolsep}{1.5pt}
    \begin{tabular}{l@{\hspace{-8pt}}cc}
        \toprule
        Setting &
        Llama-3.1-8B &
        Qwen3-8B \\
        \midrule
        Static FP16       & 49.89 & 49.59 \\
        Static FP8        & 48.72 & 49.07 \\
        \midrule
        ElasticKV (5\% cap)  & 50.06 & 49.59 \\
        ElasticKV (10\% cap) & 50.06 & 49.52 \\
        ElasticKV (20\% cap) & 50.04 & 49.42 \\
        \midrule
        All-BASE          & 48.45 & 47.87 \\
        \bottomrule
    \end{tabular}
\vspace{-3mm}
\end{wraptable}
We evaluate the generation quality on all LongBench tasks~\citep{longbench} and report the average official task-specific scores in Tab.~\ref{tab_quality}. Since BASE exposure is induced by runtime pressure rather than directly configured, we report ElasticKV under BASE-exposure caps of 5\%, 10\%, and 20\% with a tolerance of 0.5\%. For each cap, we select the valid run with the highest observed exposure within this bound for each task. Complete per-task results and exposures are reported in App.~\ref{app_quality}. We also include a manually constructed all-BASE setting, where all KV is executed in BASE, as a 100\% stress endpoint. 
This experiment calibrates the quality impact of BASE execution and establishes a quality-validated exposure range that covers the subsequent serving operating points.

ElasticKV largely preserves FP16 quality across the practical mixed-fidelity range. Under all three exposure caps, the 21-task average essentially matches FP16 on both models. The same trend holds for tasks that actually reach high exposure. For both models, 13 of the 21 tasks reach at least 18\% BASE exposure, and their subset-average scores remain within 0.1 points of FP16 (cf. App.~\ref{app_quality}). Even the All-BASE setting shows only modest degradation, while normal ElasticKV operates far from this extreme.
%

Static FP8 is a fixed-fidelity reference that keeps the target representation at 8 bits throughout execution, regardless of memory pressure. ElasticKV instead invokes 8-bit BASE only when needed and obtains higher scores than FP8. We report FP8 serving results in App.~\ref{app_fp8}, as it uses a different optimized attention backend from our runtime comparison.

\subsection{Overall End-to-End Serving Performance}
\label{sec_concurrency}

\begin{figure}[t]
\centerline{\includegraphics[width=\textwidth]{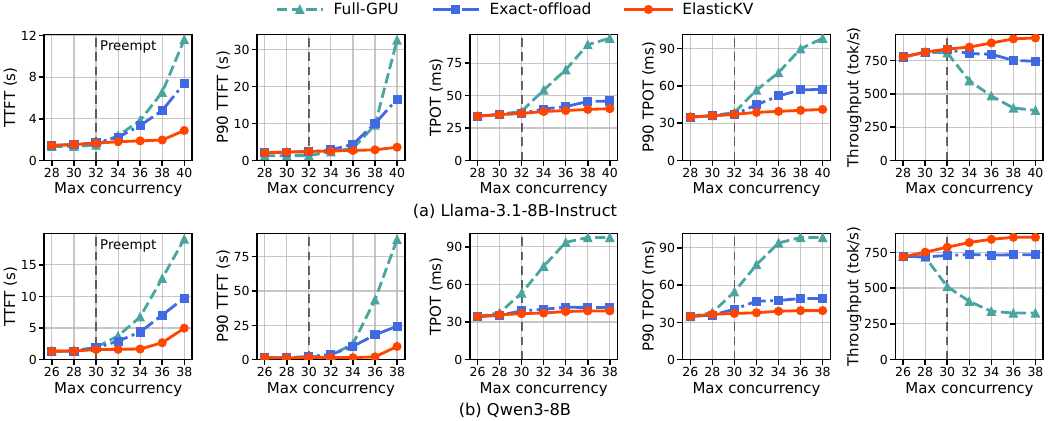}}
	\centering
    \vspace{-1mm}
	\caption{Serving performance under increasing concurrency for (a) Llama-3.1-8B-Instruct and (b) Qwen3-8B. Columns show mean/P90 TTFT, mean/P90 TPOT, and output-token throughput. Lower is better for latency; higher is better for throughput. First preemption points are also reported.}
	\label{fig_overall}
\end{figure}

We evaluate end-to-end serving performance by varying the maximum request concurrency under the synthetic workload described in \S\ref{sec_setup}. We report mean and P90 TTFT, mean and P90 TPOT, and token throughput in Fig.~\ref{fig_overall}. We use a 20GB GPU KV cache budget for both models. Increasing concurrency raises the amount of live KV cache and exposes the pressure point where the baselines begin to preempt requests. 
Overall, all three systems show comparable performance before pressure points, indicating that ElasticKV does not penalize the native TARGET-only regime. Beyond them, Full-GPU and Exact-offload exhibit a sharp latency knee, while ElasticKV remains more stable over a wider concurrency range. 

The improvement is most pronounced in TTFT, which is directly affected by preemption and recovery delay. By reclaiming GPU capacity before preemption while keeping BASE KV executable, ElasticKV extends the stable serving range beyond the baselines. For example, at a maximum concurrency of 36 on Qwen3-8B, Full-GPU and Exact-offload incur 4989 and 579 preemptions, while ElasticKV incurs none. This explains the large TTFT gap. At the highest tested concurrency, ElasticKV has 3.8-4.0$\times$ lower mean TTFT and 9.1$\times$ lower P90 TTFT than Full-GPU, and 2.0-2.6$\times$ lower mean TTFT and 2.5-4.7$\times$ lower P90 TTFT than Exact-offload, across the two models.

ElasticKV also avoids trading startup latency for unstable decoding. At the highest concurrency, Full-GPU incurs about 2.5$\times$ higher TPOT than ElasticKV, while Exact-offload shows a smaller but visible increase in tail TPOT. Meanwhile, ElasticKV obtains 1.17-1.23$\times$ the throughput of Exact-offload and 2.45-2.63$\times$ that of Full-GPU. 
These results show that ElasticKV keeps requests making progress under pressure and extends the stable serving range beyond that of exact-fidelity execution.

\subsection{Realistic Workloads and Runtime Adaptation}
\label{sec_real}

\begin{wrapfigure}{r}{.37\textwidth}
\vspace{-5mm}
\centerline{\includegraphics[width=1.0\linewidth]{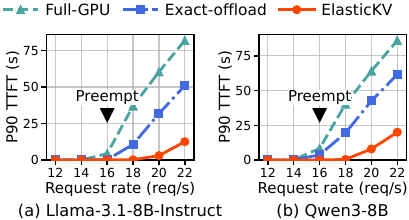}}
	\centering
    \vspace{-2mm}
	\caption{P90 TTFT on ShareGPT under increasing request rates for (a) Llama-3.1-8B-Instruct and (b) Qwen3-8B. Lower is better.}
    \label{fig_sharegpt_ttft}
\vspace{-3mm}
\end{wrapfigure}
\textbf{ShareGPT workloads.}
We next replay 5000 realistic ShareGPT~\citep{sharegpt} requests with heterogeneous request lengths and vary the arrival rate. 
As in Fig.~\ref{fig_sharegpt_ttft}, the baselines exhibit sharp P90 TTFT growth as load rises. At the highest request rate, ElasticKV achieves 4.3-6.5$\times$ lower P90 TTFT than Full-GPU, and 3.1-4.0$\times$ lower P90 TTFT than Exact-offload. Both baselines exceed 1000 preemptions on each model, while ElasticKV remains preemption-free with only 2.6\% and 3.6\% BASE exposure. Additional metrics are provided in App.~\ref{app_sharegpt}.

\begin{figure}[t]
\centerline{\includegraphics[width=\textwidth]{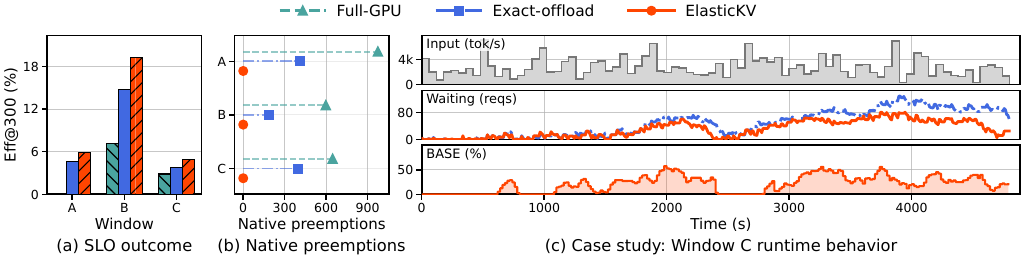}}
	\centering
    \vspace{-1mm}
	\caption{Mooncake production trace replay with $d=16$. (a) Eff@300 across three trace windows, higher is better; (b) native request preemptions, lower is better; (c) Window C runtime behavior showing offered input load, waiting requests, and ElasticKV's resident BASE ratio over time.}
	\label{fig_mooncake_a100}
\end{figure}

\noindent\textbf{Production trace replay.}
We further evaluate ElasticKV using the production trace released with Mooncake~\citep{mooncake}. We replay three 300s windows with different temporal load patterns. To adapt the production traffic to our single-GPU testbed, we dilate arrival times by $d=16$. 
Following Mooncake, we report effective request ratio Eff@300, the fraction of requests with TTFT $<30$s and time-between-token (TBT) $<300$ms. Trace selection and metric definitions are in App.~\ref{app_mooncake}.

As shown in Fig.~\ref{fig_mooncake_a100}, ElasticKV improves Eff@300 across all three windows while remaining preemption-free.
In the highest-load window C, Exact-offload develops a larger and more persistent waiting queue, while ElasticKV reduces the mean waiting by 36\%. The intermittent BASE residency shows that reduced fidelity is invoked on demand rather than used continuously in ElasticKV. A lighter $d=20$ trace replay shows the
\begin{wrapfigure}{r}{.53\textwidth}
\vspace{2mm}
\centerline{\includegraphics[width=1.0\linewidth]{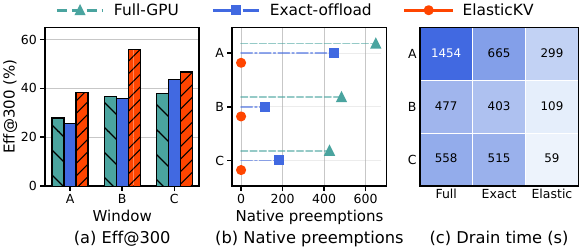}}
	\centering
    \vspace{-1mm}
	\caption{Production-trace replay with Llama-3.1-70B-Instruct on 8$\times$ AMD MI355X  (TP2$\times$DP4). (a) Eff@300, higher is better; (b) native request preemptions, lower is better; and (c) drain time, lower is better.}
    \label{fig_mi355x}
\vspace{-14mm}
\end{wrapfigure}
same trend and is reported in App.~\ref{app_mooncake}.

\subsection{Large-Scale and Cross-Platform Generalization}
\label{sec_amd}

To evaluate ElasticKV beyond the single-A100 setting, we deploy Llama-3.1-70B-Instruct on eight AMD MI355X GPUs using 2-way tensor parallelism (TP) and 4-way data parallelism (DP), denoted TP2$\times$DP4. We replay three 600s Mooncake windows with $d=5$,  jointly varying accelerator platform, model scale, and distributed topology. Detailed configurations are provided in App.~\ref{app_mi355x}.

Figure~\ref{fig_mi355x} shows that ElasticKV retains and even strengthens its advantages at this larger scale. Across three windows, ElasticKV improves average Eff@300 by 34.1\% over Exact-offload, eliminates native request preemptions, and reduces average drain time by 70.5\%. The gains over Full-GPU are larger. These results confirm that our mixed-fidelity execution remains effective with a substantially larger model, combined DP and TP execution, and a different accelerator platform. Results for the TP1$\times$DP8 topology and complete metrics are reported in App.~\ref{app_mi355x}.


\subsection{Sensitivity Analysis}
\label{sec_sensitivity}

\begin{wrapfigure}{r}{.4\textwidth}
\vspace{-5mm}
\centerline{\includegraphics[width=1.0\linewidth]{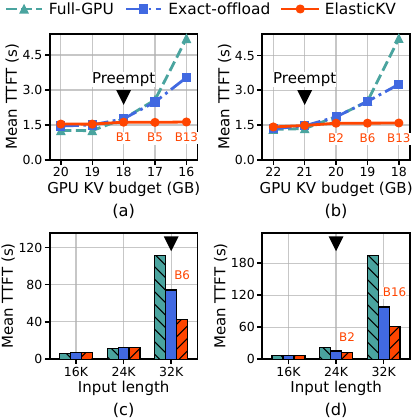}}
	\centering
    \vspace{-2mm}
	\caption{Mean TTFT under varying GPU KV budgets (a, b) and input context lengths (c, d) for Llama-3.1-8B-Instruct (a, c) and Qwen3-8B (b, d). Lower is better. ``B'' denotes ElasticKV's BASE exposure (\%).}
    \label{fig_sensitivity}
\vspace{-10mm}
\end{wrapfigure}
We next vary two additional sources of KV pressure: GPU KV capacity and input context length.


\textbf{GPU KV budget.} Figure~\ref{fig_sensitivity}(a, b) compares mean TTFT across different GPU KV budgets at a fixed maximum concurrency of 30. When the KV budget is sufficient, all systems show comparable TTFT. As the budget tightens, the baselines cross their preemption boundaries and TTFT rises sharply, while ElasticKV remains more stable through BASE execution. Meanwhile, ElasticKV's BASE exposure remains at only 13\% on both models at the tightest tested budgets.

\textbf{Input context length.} Figure~\ref{fig_sensitivity}(c, d) varies the input length with maximum concurrency fixed at 16. As longer contexts push the system across the KV-capacity boundary, the baselines begin to preempt requests, while ElasticKV remains preemption-free. At 32K input length, ElasticKV reduces mean TTFT over Exact-offload by 21.7\% and 21.8\% on Llama-3.1-8B-Instruct and Qwen3-8B, with only 6.3\% and 15.9\% BASE exposure, respectively.

\subsection{System Analysis}
\label{sec_overhead}

\begin{wraptable}{r}{0.28\textwidth}
    \vspace{-6mm}
    \small
    \centering
    \caption{ElasticKV-specific cost under high pressure. }
    \vspace{-2mm}
    \label{tab_overhead}
    \setlength{\tabcolsep}{2pt}
    \begin{tabular}{lc}
        \toprule
        Activity & Cost (ms/step) \\
        \midrule
        Online control     & 0.32 \\
        Sync. demotion     & 2.82 \\
        Elastic metadata   & 1.23 \\
        \bottomrule
    \end{tabular}
    \vspace{-6mm}
\end{wraptable}
\textbf{Runtime overhead.} We profile the Llama-3.1-8B-Instruct setting from \S\ref{sec_concurrency} at maximum concurrency 38. As shown in Tab.~\ref{tab_overhead}, online control and elastic metadata add 0.32ms and 1.23ms/step. Synchronous demotion is invoked on 11.1\% of steps and contributes 2.82 ms/step when averaged over the full run. Overall, ElasticKV adds 4.37 ms/step of measured critical-path work. Promotion proceeds asynchronously and is excluded from this total. Detailed profiling results are provided in App.~\ref{app_overhead}.

\begin{wraptable}{r}{0.42\textwidth}
    \vspace{-4mm}
    \small
    \centering
    \caption{Ablation under high pressure. Lower is better for P90 TTFT and P90 TPOT. BASE denotes exposure.}
    \vspace{-1mm}
    \label{tab_ablation}
    \setlength{\tabcolsep}{2pt}
    \begin{tabular}{l@{\hspace{-6pt}}ccc}
        \toprule
        Variant &
        P90 TTFT &
        P90 TPOT &
        BASE \\
        \midrule
        ElasticKV            & 2.8s & 40.3ms & 14.7\% \\
        Watermark demotion   & 3.4s & 46.5ms & 17.2\% \\
        Eager promotion      & 4.4s & 52.0ms & 15.7\% \\
        No promotion         & 2.6s & 40.5ms & 28.8\% \\
        General elastic path & 3.1s & 49.5ms & 14.7\% \\
        \bottomrule
    \end{tabular}
    \vspace{-1mm}
\end{wraptable}
\textbf{Ablation study.}
Under the same setting, Tab.~\ref{tab_ablation} isolates the main runtime mechanisms. Watermark demotion increases both tail latency and BASE exposure. Eager promotion causes the largest TTFT penalty, raising P90 TTFT from 2.8s to 4.4s. Conversely, disabling promotion slightly improves P90 TTFT but nearly doubles BASE exposure, showing its role in fidelity recovery. Finally, forcing all mixed-fidelity batches through the general elastic path increases P90 TPOT without changing BASE exposure, validating the benefit of structured attention dispatch. Detailed variant definitions are reported in App.~\ref{app_ablation}.

\section{Conclusion}

We present ElasticKV, a mixed-fidelity KV runtime that relaxes the target-only execution contract in LLM serving by allowing requests to make progress with an intermediate executable state under memory pressure. ElasticKV combines a pair-structured KV layout, a dual-mode attention backend, and pressure-aware fidelity control to make this state physically reclaimable, directly executable, and dynamically managed at runtime. Our evaluation shows that ElasticKV preserves generation quality while substantially reducing latency across workloads, models, and GPU platforms. More broadly, ElasticKV suggests that KV cache optimization should treat fidelity not only as a representation choice, but as a runtime-managed execution property that can adapt to resource pressure.

\newpage
\bibliography{main}
\newpage
\appendix
\section*{Appendix}
\section{LLM Inference and KV-Cache Lifecycle}
\label{app_bg}

This section supplements the paged-KV background in \S\ref{sec_paged} with additional details on how KV state is created, grows, and is reclaimed over the lifetime of a serving request.

An LLM inference request typically consists of a \textit{prefill} phase followed by autoregressive \textit{decoding}. During prefill, the model processes the input prompt and materializes the corresponding key-value (KV) states for attention. During decoding, tokens are generated autoregressively, and each newly generated token appends new KV states that are reused by subsequent attention computation. Maintaining this states avoids repeatedly recomputing the attention states of previously processed tokens, but also causes the memory footprint of each active request to grow as the sequence progresses. 

The amount of live KV memory therefore changes dynamically during serving. Concurrent requests may have different arrival times, context lengths, and generation lengths, while completed or preempted requests release their cached states. Under high concurrency or long contexts, this aggregate live footprint can approach the available GPU KV capacity, making KV allocation a direct constraint on whether requests can continue to make progress.

Modern systems commonly manage this changing footprint using paged KV cache management~\citep{vllm}. Rather than requiring a contiguous KV region for each request, the runtime maps logical KV blocks to fixed-size physical blocks and allocates or reclaims these blocks as requests progress. This abstraction reduces external fragmentation and enables block-granular movement and reuse, but physical capacity is still reclaimed at the page granularity.

\section{Runtime Implementation Details}
\label{app_impl}

This section supplements the implementation overview in \S\ref{sec_impl} with lower-level details on state tracking, transition protocols, scheduler integration, and elastic attention execution. Following the three integration points, we describe the state and scheduler integration, transition engine, and elastic attention backend.

\subsection{State and Scheduler Integration}

\paragraph{Pair-state tracking.}
ElasticKV augments \texttt{KVCacheManager} with a \texttt{LogicalBlockStateTable}. Each physical pair is represented by an \texttt{ElasticPairEntry}. The table is the source of truth for transition eligibility and physical mapping, while the original logical block namespace remains unchanged. Thus, the serving stack continue to operate on standard logical KV blocks, even when their physical representation is managed as a pair.

The state table exposes queries for demotable TARGET pairs and promotable BASE pairs rather than allowing the scheduler to manipulate physical blocks directly. Demotion eligibility excludes pairs without recoverable CPU backing and pairs protected by the runtime, including attention-sink and recent-token regions. Among the remaining candidates, ElasticKV uses recent use and request status to prioritize non-urgent pairs. The selection policy remains intentionally simple. The objective is to recover the required capacity rather than solve a global KV-selection problem.

\paragraph{Scheduler hooks.}
ElasticKV leaves vLLM's native request and token scheduling policy unchanged and inserts three lightweight hooks around KV allocation and transition management. 
\begin{itemize}[leftmargin=*,itemsep=2pt,topsep=3pt,parsep=0pt]
\item \texttt{probe\_shortage} performs the side-effect-free demand probe. To avoid running the probe when memory pressure is clearly absent, ElasticKV uses a static capacity guard $A_{\max}=\max(N_{\mathrm{req}},\lceil N_{\mathrm{tok}}/B\rceil)$, where $N_{\mathrm{req}}$ and $N_{\mathrm{tok}}$ are the maximum numbers of requests and tokens that can advance in one scheduling step, and $B$ is the KV block size. These quantities are fixed by the configuration, so $A_{\max}$ is computed once at initialization. When the number of free blocks is above $A_{\max}$, the shortage probe is skipped.

\item \texttt{emit\_demotion} submits the selected TARGET pairs to the transition engine when a positive shortage is observed, and commits the recovered capacity only after the transition completes.

\item \texttt{emit\_promotion} uses the reserve rule after required scheduling and allocation, and asynchronously issues eligible BASE-to-TARGET refinements. BASE-ready requests remain execution-ready throughout this process and do not wait for the promotion.
\end{itemize}

\paragraph{Default control parameters.}
Our default configuration uses a minimum demotion batch $B^d_{\min}=16$ pairs and a reserve decay of $\lambda=2$ blocks per
scheduling step. These values are chosen from profiling the exact on-demand shortage behavior. Individual shortages are typically small, with a common shortage of roughly two blocks. Setting $B^d_{\min}=16$ amortizes GPU transition and metadata overhead when such small shortages occur repeatedly. The reserve decay $\lambda=2$ then releases surplus capacity back to the promotion budget at approximately the rate of near-term allocation demand. Both hyperparameters are runtime-control defaults and workload-agnostic.

\subsection{Transition Protocols}

ElasticKV implements fidelity transitions in a worker-side \texttt{ElasticTransitionEngine}. A TARGET-to-BASE transition is synchronous. For each selected pair, the GPU packs the high-order bytes of the two TARGET blocks into the pair's BASE layout described in \S\ref{sec_layout}. The physical block released by this packing is not returned to the allocator until the GPU transition has completed and the new mapping is committed. 

BASE-to-TARGET promotion follows the opposite resource dependency and is asynchronous. Promotion first reserves the additional GPU block required by the TARGET pair and issues restoration from CPU backing. The existing BASE mapping remains valid and executable while the transfer is in flight. ElasticKV tracks completion events and publishes the mapping only after TARGET has been materialized. Consequently, promotion does not block the current request progress. 

The current implementation performs promotion in one step from the 8-bit BASE state to the original 16-bit TARGET state. BASE stores the high-order byte of each original TARGET, so we do not introduce a separate (de)compression process.
For tensor-parallel execution, fidelity transitions are coordinated across the participating TP ranks. A pair transition is committed consistently across ranks so that workers do not observe different fidelity states for the same logical pair.

\subsection{Elastic Attention Backend}
\label{app_impl_attn}

ElasticKV registers an \texttt{ELASTIC\_TRITON\_ATTN} backend with the vLLM model runner. The runtime communicates mixed-fidelity state using sparse per-request metadata rather than replacing the native attention interface. The worker lowers this metadata into GPU-side execution structures, including a batch-level BASE indicator \texttt{has\_base\_seq}, an elastic physical block table \texttt{elastic\_block\_table}, compact boundary metadata \texttt{elastic\_boundaries} for structured mode sequences, and an explicit \texttt{pair\_mode\_table} for irregular sequences.

The batch-level BASE indicator provides the first dispatch decision. TARGET-only batches invoke the native Triton attention path. For mixed-fidelity batches, the worker first inspects the mode layout. Sequences with the common structured TARGET-BASE-TARGET pattern are encoded using compact mode boundaries and dispatched to the structured elastic path. Irregular layouts use the explicit pair-mode table and the general elastic path. This keeps the general mixed-fidelity machinery off the native path and avoids dense per-pair mode lookup for regular mixed layouts.

TARGET and BASE pairs share the same attention output interface but use different load semantics. TARGET pairs follow the native 16-bit load path. For a BASE value, the kernel loads its stored high-order byte, shifts it into the high-order position of the native 16-bit representation, and bitcasts the result to the attention datatype. The unavailable low-order byte is therefore zero-filled in registers. This operation requires neither a quantization scale nor a separate dequantization transform, and no TARGET is materialized before BASE attention executes. Newly appended KV is written in TARGET form and enters the protected recent region before becoming eligible for later demotion.

\begin{table}[ht!]
\centering
\caption{Per-task LongBench generation-quality results for Llama-3.1-8B-Instruct. For each ElasticKV exposure cap, we report the official task score and the observed BASE exposure (\textit{Exp.}) of the selected run. Higher scores are better for all task metrics.}
\label{tab_longbench_llama}
\scriptsize
\setlength{\tabcolsep}{2.4pt}
\resizebox{\textwidth}{!}{
\begin{tabular}{llrrrrrrrrr}
\toprule
& &
\multicolumn{3}{c}{Reference settings} &
\multicolumn{6}{c}{ElasticKV exposure cap} \\
\cmidrule(lr){3-5}
\cmidrule(lr){6-11}
Task & Metric &
FP16 &
FP8 &
All-BASE &
\multicolumn{2}{c}{$\leq 5\%$} &
\multicolumn{2}{c}{$\leq 10\%$} &
\multicolumn{2}{c}{$\leq 20\%$} \\
\cmidrule(lr){6-7}
\cmidrule(lr){8-9}
\cmidrule(lr){10-11}
& & & & &
Score & Exp. &
Score & Exp. &
Score & Exp. \\
\midrule

\multicolumn{11}{l}{\textit{Single-Document QA}} \\
NarrativeQA
    & QA F1            & 27.85 & 28.34 & 27.61
    & 28.87 & 5.3\% & 28.95 & 6.8\% & 28.95 & 6.8\% \\
Qasper
    & QA F1            & 44.76 & 43.81 & 44.88
    & 44.83 & 5.2\% & 44.69 & 8.6\% & 44.61 & 20.5\% \\
MultiFieldQA-en
    & QA F1            & 55.62 & 53.38 & 55.48
    & 56.01 & 3.7\% & 56.09 & 10.5\% & 56.08 & 20.5\% \\
MultiFieldQA-zh
    & QA F1 (zh)       & 63.47 & 61.35 & 60.44
    & 63.61 & 4.8\% & 63.54 & 9.8\% & 63.30 & 16.5\% \\

\addlinespace[2pt]
\multicolumn{11}{l}{\textit{Multi-Document QA}} \\
HotpotQA
    & QA F1            & 58.62 & 56.71 & 54.44
    & 58.04 & 4.8\% & 57.95 & 7.2\% & 57.95 & 7.2\% \\
2WikiMQA
    & QA F1            & 48.85 & 46.76 & 47.89
    & 49.25 & 5.1\% & 49.65 & 9.1\% & 49.65 & 9.1\% \\
MuSiQue
    & QA F1            & 32.72 & 28.83 & 29.36
    & 32.11 & 4.9\% & 31.1 & 10.2\% & 31.04 & 18.1\% \\
DuReader
    & ROUGE-L (zh)     & 34.95 & 32.34 & 33.17
    & 33.87 & 4.6\% & 34.67 & 10.5\% & 34.90 & 19.9\% \\

\addlinespace[2pt]
\multicolumn{11}{l}{\textit{Summarization}} \\
GovReport
    & ROUGE-L          & 34.62 & 34.58 & 33.87
    & 34.55 & 4.8\% & 34.65 & 9.6\% & 34.21 & 19.7\% \\
QMSum
    & ROUGE-L          & 25.21 & 25.25 & 25.64
    & 25.45 & 5.1\% & 25.36 & 10.5\% & 25.21 & 13.1\% \\
MultiNews
    & ROUGE-L          & 26.81 & 26.66 & 26.17
    & 26.86 & 4.9\% & 26.82 & 9.3\% & 26.98 & 20.5\% \\
VCSum
    & ROUGE-L (zh)     & 17.22 & 17.53 & 17.32
    & 17.39 & 5.3\% & 17.30 & 10.4\% & 17.17 & 18.1\% \\

\addlinespace[2pt]
\multicolumn{11}{l}{\textit{Few-Shot Learning}} \\
TREC
    & Classification Acc. & 73.00 & 71.50 & 73.00
    & 73.00 & 5.5\% & 73.00 & 9.7\% & 73.00 & 20.5\% \\
TriviaQA
    & QA F1            & 91.65 & 90.22 & 90.90
    & 91.65 & 4.8\% & 91.65 & 9.2\% & 91.65 & 20.5\% \\
SAMSum
    & ROUGE-L          & 43.67 & 44.44 & 44.38
    & 43.70 & 5.0\% & 43.88 & 10.4\% & 43.86 & 20.5\% \\
LSHT
    & Classification Acc. & 46.00 & 46.25 & 44.00
    & 46.50 & 5.3\% & 46.50 & 8.6\% & 47.00 & 18.3\% \\

\addlinespace[2pt]
\multicolumn{11}{l}{\textit{Synthetic Tasks}} \\
PassageCount
    & Count Acc.       & 8.58 & 6.82 & 5.06
    & 10.08 & 4.3\% & 10.08 & 4.3\% & 10.08 & 4.3\% \\
PassageRetrieval-en
    & Retrieval Acc.   & 99.50 & 99.50 & 98.50
    & 99.50 & 0 & 99.50 & 0 & 99.50 & 0 \\
PassageRetrieval-zh
    & Retrieval Acc. (zh) & 90.70 & 86.85 & 90.83
    & 92.02 & 4.1\% & 92.02 & 4.1\% & 92.02 & 4.1\% \\

\addlinespace[2pt]
\multicolumn{11}{l}{\textit{Code Completion}} \\
LCC
    & Code Sim.        & 65.16 & 63.27 & 60.60
    & 65.20 & 4.6\% & 65.10 & 10.5\% & 65.09 & 20.5\% \\
RepoBench-P
    & Code Sim.        & 58.76 & 58.69 & 53.95
    & 58.83 & 5.3\% & 58.78 & 10.5\% & 58.59 & 19.6\% \\

\midrule
\textbf{Average}
    & --
    & \textbf{49.89}
    & \textbf{48.72}
    & \textbf{48.45}
    & \textbf{50.06} & --
    & \textbf{50.06} & --
    & \textbf{50.04} & -- \\
\bottomrule
\end{tabular}
}
\end{table}

\begin{table}[ht!]
\centering
\caption{Per-task LongBench generation-quality results for Qwen3-8B. For each ElasticKV exposure cap, we report the official task score and the observed BASE exposure (\textit{Exp.}) of the selected run. Higher scores are better for all task metrics.}
\label{tab_longbench_qwen}
\scriptsize
\setlength{\tabcolsep}{2.4pt}
\resizebox{\textwidth}{!}{
\begin{tabular}{llrrrrrrrrr}
\toprule
& &
\multicolumn{3}{c}{Reference settings} &
\multicolumn{6}{c}{ElasticKV exposure cap} \\
\cmidrule(lr){3-5}
\cmidrule(lr){6-11}
Task & Metric &
FP16 &
FP8 &
All-BASE &
\multicolumn{2}{c}{$\leq 5\%$} &
\multicolumn{2}{c}{$\leq 10\%$} &
\multicolumn{2}{c}{$\leq 20\%$} \\
\cmidrule(lr){6-7}
\cmidrule(lr){8-9}
\cmidrule(lr){10-11}
& & & & &
Score & Exp. &
Score & Exp. &
Score & Exp. \\
\midrule

\multicolumn{11}{l}{\textit{Single-Document QA}} \\
NarrativeQA
    & QA F1            & 26.32 & 24.39 & 24.73
    & 26.88 & 4.8\% & 26.64 & 6.7\% & 26.64 & 6.7\% \\
Qasper
    & QA F1            & 47.70 & 45.87 & 45.08
    & 47.36 & 5.5\% & 47.03 & 10.3\% & 47.4 & 18.0\% \\
MultiFieldQA-en
    & QA F1            & 53.34 & 54.34 & 51.33
    & 52.83 & 3.9\% & 52.93 & 10.5\% & 52.56 & 19.6\% \\
MultiFieldQA-zh
    & QA F1 (zh)       & 63.07 & 63.03 & 62.66
    & 62.91 & 5.0\% & 63.11 & 8.0\% & 63.11 & 8.0\% \\

\addlinespace[2pt]
\multicolumn{11}{l}{\textit{Multi-Document QA}} \\
HotpotQA
    & QA F1            & 59.23 & 58.27 & 56.57
    & 59.03 & 4.9\% & 58.98 & 7.4\% & 58.98 & 7.4\% \\
2WikiMQA
    & QA F1            & 43.37 & 43.18 & 40.11
    & 42.87 & 1.2\% & 42.87 & 1.2\% & 42.87 & 1.2\% \\
MuSiQue
    & QA F1            & 36.30 & 31.61 & 29.67
    & 36.15 & 4.3\% & 36.55 & 10.5\% & 34.26 & 17.9\% \\
DuReader
    & ROUGE-L (zh)     & 27.09 & 27.21 & 25.85
    & 26.42 & 4.5\% & 25.80 & 9.8\% & 25.95 & 19.2\% \\

\addlinespace[2pt]
\multicolumn{11}{l}{\textit{Summarization}} \\
GovReport
    & ROUGE-L          & 33.63 & 33.12 & 32.02
    & 33.71 & 3.6\% & 33.35 & 8.8\% & 33.33 & 20.5\% \\
QMSum
    & ROUGE-L          & 24.11 & 23.64 & 23.22
    & 24.35 & 4.5\% & 24.50 & 10.1\% & 24.50 & 12.1\% \\
MultiNews
    & ROUGE-L          & 24.84 & 24.69 & 24.42
    & 24.69 & 4.1\% & 24.65 & 6.5\% & 24.34 & 20.5\% \\
VCSum
    & ROUGE-L (zh)     & 14.01 & 14.10 & 14.86
    & 14.12 & 4.0\% & 14.20 & 8.6\% & 14.14 & 20.5\% \\

\addlinespace[2pt]
\multicolumn{11}{l}{\textit{Few-Shot Learning}} \\
TREC
    & Classification Acc. & 71.50 & 72.00 & 73.00
    & 72.00 & 5.1\% & 72.50 & 7.6\% & 72.50 & 20.5\% \\
TriviaQA
    & QA F1            & 90.21 & 89.78 & 90.03
    & 90.21 & 5.5\% & 90.61 & 10.5\% & 90.61 & 20.0\% \\
SAMSum
    & ROUGE-L          & 44.59 & 45.38 & 44.33
    & 44.49 & 5.1\% & 44.16 & 10.5\% & 44.17 & 20.5\% \\
LSHT
    & Classification Acc. & 47.50 & 46.50 & 45.75
    & 49.25 & 5.0\% & 48.50 & 9.0\% & 47.50 & 19.7\% \\

\addlinespace[2pt]
\multicolumn{11}{l}{\textit{Synthetic Tasks}} \\
PassageCount
    & Count Acc.       & 1.50 & 2.00 & 4.75
    & 1.50 & 5.2\% & 1.50 & 10.5\% & 2.50 & 20.5\% \\
PassageRetrieval-en
    & Retrieval Acc.   & 100.00 & 99.67 & 92.00
    & 99.50 & 5.0\% & 99.00 & 10.0\% & 99.50 & 10.0\% \\
PassageRetrieval-zh
    & Retrieval Acc. (zh) & 98.00 & 97.33 & 96.50
    & 98.00 & 0 & 98.00 & 0 & 98.00 & 0 \\

\addlinespace[2pt]
\multicolumn{11}{l}{\textit{Code Completion}} \\
LCC
    & Code Sim.        & 69.33 & 68.69 & 66.56
    & 69.43 & 4.7\% & 69.30 & 9.0\% & 69.53 & 20.5\% \\
RepoBench-P
    & Code Sim.        & 65.71 & 65.69 & 61.77
    & 65.62 & 4.8\% & 65.66 & 10.5\% & 65.33 & 20.5\% \\

\midrule
\textbf{Average}
    & --
    & \textbf{49.59}
    & \textbf{49.07}
    & \textbf{47.87}
    & \textbf{49.59} & --
    & \textbf{49.52} & --
    & \textbf{49.42} & -- \\
\bottomrule
\end{tabular}
}
\end{table}

\section{Additional Experiments}
\label{app_exp}

\subsection{Full LongBench Quality Results}
\label{app_quality}

Our quality evaluation in \S\ref{sec_quality} focuses on task-level generation quality under KV fidelity changes, while broader model reliability under perturbations and distribution shifts has been extensively studied in prior work~\citep{madry2017towards, hendrycks2019benchmarking, cohen2019certified, yang2026semantic, ilyas2019adversarial, yang2024regulating, ovadia2019can, mitchell2021fast, yang2026attribution}.
This section supplements the aggregate evaluation in \S\ref{sec_quality} with the complete per-task LongBench~\citep{longbench} results and the observed BASE exposure of each selected ElasticKV run. We evaluate all 21 tasks using their official task-specific metrics and retain the same task set for every reported average.
We define BASE exposure as the fraction of attention-visible KV pairs executed in BASE, aggregated over all execution steps. It is induced by runtime pressure rather than directly configured, and different tasks therefore need not reach exactly the same exposure level. We use exposure caps $c\in\{5\%,10\%,20\%\}$ with a tolerance of 0.5 percentage points to accommodate runtime deviations. For each task and cap, we select the valid completed run with the highest observed BASE exposure satisfying $e(r)\leq c+0.5\%$. Selection depends only on run validity and achieved exposure. If a task does not reach a higher exposure within a cap, the same run may be selected under multiple caps.

Tables~\ref{tab_longbench_llama} and~\ref{tab_longbench_qwen} report both the official task score and the actual BASE exposure of each selected ElasticKV run. We additionally report FP16, FP8, and All-BASE stress endpoint as reference settings.
The cap in ElasticKV is an upper bound rather than a target exposure. Because BASE is invoked only in response to runtime memory pressure, the achieved exposure varies naturally across tasks. Some selected runs naturally remain well below the cap or execute entirely in TARGET. A 0\% exposure denotes a run where BASE was never exercised, rather than missing data. Since we select only observed operating points, a low exposure can also occur when the next more aggressive configuration exceeds the cap. We retain such cases instead of forcing BASE execution, preserving the same task set and selection rule across all caps and both models.

Across all 21 tasks, ElasticKV remains close to FP16 throughout the practical mixed-fidelity regime. Relative to FP16, the average-score differences under the 5\%, 10\%, and 20\% caps are $+0.17$, $+0.17$, and $+0.15$ points for Llama-3.1-8B-Instruct, and $0.00$, $-0.07$, and $-0.17$ points for Qwen3-8B. These small non-monotonic differences should not be interpreted as quality improvements. They indicate no systematic degradation under the exposure regimes. In contrast, forcing all KV to execute in BASE lowers the average by 1.44 points on Llama-3.1-8B-Instruct and 1.72 points on Qwen3-8B. This separation between the practical cap regimes and the All-BASE endpoint supports using BASE adaptively under pressure rather than as a permanent global fidelity setting.

\begin{figure}[t]
\centerline{\includegraphics[width=0.8\textwidth]{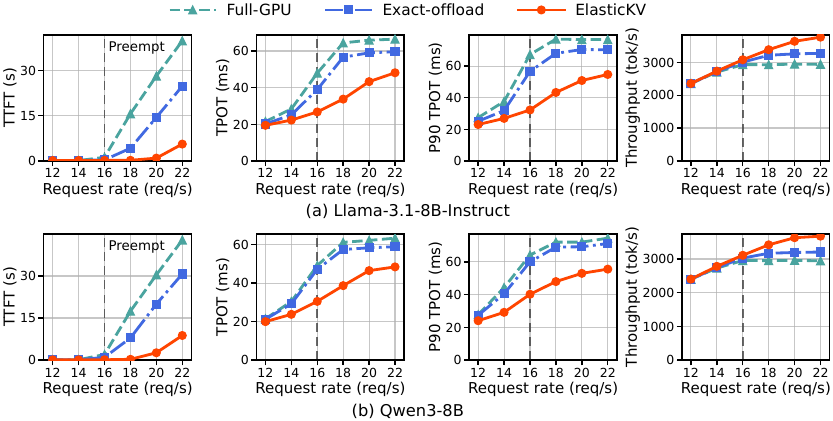}}
	\centering
	\caption{Serving performance on ShareGPT under increasing request rates for (a) Llama-3.1-8B-Instruct and (b) Qwen3-8B. Columns show mean TTFT, mean/P90 TPOT, and output-token throughput. Lower is better for latency; higher is better for throughput. First preemptions are also reported.}
	\label{fig_sharegpt_all}
\end{figure}

Since cap-level averages also contain tasks that naturally remain at low exposure, we additionally examine a near-cap subset whose selected run under the 20\% cap reaches at least 18\% BASE exposure. For both models, 13/21 tasks meet this criterion. Their average scores are 48.01 and 45.37 for Llama-3.1-8B-Instruct and Qwen3-8B, compared with 48.07 and 45.46 under FP16, respectively. Thus, both subset averages remain within 0.1 points of FP16, showing that the near-FP16 aggregate quality is not merely an artifact of tasks remaining at low BASE exposure.

\subsection{Additional ShareGPT Results}
\label{app_sharegpt}

This section provides complementary metrics for the ShareGPT~\citep{sharegpt} evaluation in
\S\ref{sec_real}. 
Figure~\ref{fig_sharegpt_all} extends the P90 TTFT results in Fig.~\ref{fig_sharegpt_ttft} with mean TTFT, mean and P90 TPOT, and output-token throughput. At lower request rates, three systems remain relatively close. As load increases and the baselines cross their preemption boundaries, their latency rises sharply, whereas ElasticKV maintains substantially more stable latency. ElasticKV also sustains higher output throughput, showing that its latency improvement does not come from reducing serving progress.

Table~\ref{tab_sharegpt_runtime} reports runtime behavior at the highest tested request rate of 22 requests/s. Full-GPU and Exact-offload incur 1297 and 1392 native request preemptions on Llama-3.1-8B-Instruct, and 1574 and 1594 on Qwen3-8B, respectively. ElasticKV remains preemption-free on both models while using only 2.6\% and 3.6\% BASE exposure. Therefore, a small amount of pressure-triggered BASE execution is sufficient to avoid the much larger disruption caused by request preemption under this heterogeneous workload.

\begin{table}[t]
\centering
\small
\caption{Runtime behavior at the highest ShareGPT request rate
(i.e., 22 requests/s).}
\label{tab_sharegpt_runtime}
\setlength{\tabcolsep}{4pt}
\begin{tabular}{llcc}
\toprule
Model & System &
Preemptions & BASE exposure \\
\midrule
\multirow{3}{*}{Llama-3.1-8B-Instruct}
 & Full-GPU      & 1297 & -- \\
 & Exact-offload & 1392 & -- \\
 & ElasticKV     & 0    & 2.6\% \\
\midrule
\multirow{3}{*}{Qwen3-8B}
 & Full-GPU      & 1574 & -- \\
 & Exact-offload & 1594 & -- \\
 & ElasticKV     & 0    & 3.6\% \\
\bottomrule
\end{tabular}
\end{table}

\subsection{Mooncake Trace Replay and Additional Results}
\label{app_mooncake}

\subsubsection{Trace and Replay Protocol}

In \S\ref{sec_real}, we reply the conversation trace released with Mooncake~\citep{mooncake} using Llama-3.1-8B-Instruct on the single-A100 testbed. The trace contains timestamped production requests with input and output lengths. We select three temporally separated 300s windows: A=$[450,750)$, B=$[1650,1950)$, and C=$[2850,3150)$ seconds, summarized in Tab.~\ref{tab_mooncake_trace}. This samples distinct temporal regions of the trace rather than concentrating the evaluation on a single load pattern. We keep the original request order, input/output lengths, and relative arrival pattern without reordering.

To adapt the production arrival rate to our single-GPU testbed, we scale every inter-arrival interval by a dilation factor $d$. Thus, a 300s trace window is replayed over approximately 4800s with $d=16$ and 6000s with $d=20$. A larger $d$ therefore corresponds to a lighter offered load while preserving the relative arrival pattern. As shown in Tab.~\ref{tab_mooncake_trace}, window C has the highest offered token load under both replay rates and is therefore used for the runtime case study in Fig.~\ref{fig_mooncake_a100}(c).

\begin{table}[t]
\centering
\small
\caption{Mooncake conversation-trace windows used for the single-A100 replay. Token counts are totals over the original 300s trace intervals before arrival-time dilation.}
\label{tab_mooncake_trace}
\begin{tabular}{llrrrr}
\toprule
Window & Region & Trace interval & Requests & Input tokens & Output tokens \\
\midrule
A & Early  & [450,750)   & 920   & 11.82M & 0.324M \\
B & Middle & [1650,1950) & 1027  & 11.54M & 0.357M \\
C & Late   & [2850,3150) & 1122  & 13.37M & 0.358M \\
\bottomrule
\end{tabular}
\end{table}

\subsubsection{Mooncake Metrics}

\paragraph{Effective-request ratio.}
Following Mooncake~\citep{mooncake}, we compute the time-between-token (TBT) of each request as the mean of its slowest 10\% consecutive inter-token intervals. For a request with $n$ output tokens, let $\{\delta_i\}_{i=1}^{n-1}$ denote the observed inter-token intervals and $k=\max(1,\lceil0.1(n-1)\rceil)$. TBT is the mean of the $k$ largest $\delta_i$ values. A request is considered effective under threshold $\tau$ if
\[
    \mathrm{TTFT} < 30\,\mathrm{s}
    \quad\text{and}\quad
    \mathrm{TBT} < \tau .
\]
We therefore compute
\[
    \mathrm{Eff@\tau}
    =
    \frac{
    \#\{r:\mathrm{TTFT}_r<30\,\mathrm{s}
    \land \mathrm{TBT}_r<\tau\}
    }{N_{\mathrm{requests}}},
\]
and use $\tau=300$ms throughout the Mooncake evaluation.

\paragraph{Runtime statistics.}
The runtime case studies additionally report offered input load, waiting-queue length, and resident BASE ratio. Drain time is the elapsed time from the injection of the final trace request until all requests in the replay have completed. Offered input load denotes the input-token volume carried by arriving requests per unit replay time. Waiting-queue statistics are derived from periodic server-pressure samples over the replay and drain period. For ElasticKV, resident BASE ratio is obtained from lifecycle samples collected every 5s. We use resident BASE ratio only to visualize runtime adaptation and do not interpret sampled BASE episodes as demotion events. It is an instantaneous residency statistic and is distinct from the cumulative BASE exposure reported in the experiments.

\subsubsection{Lighter Production Trace Replay}

\begin{figure}[t]
\centerline{\includegraphics[width=\textwidth]{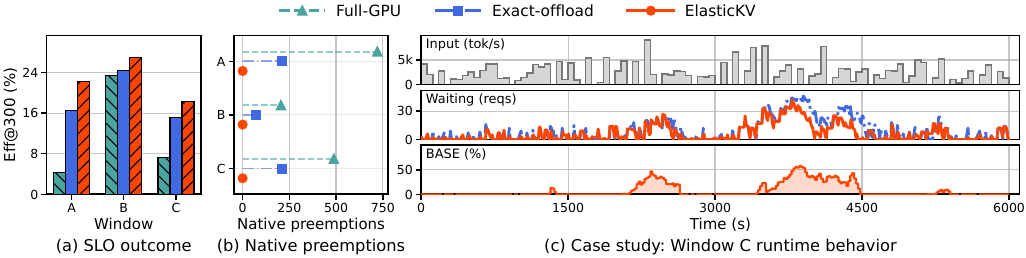}}
	\centering
	\caption{Mooncake production trace replay with $d=20$. (a) Eff@300 across three trace windows. (b) Number of native request preemptions. (c) Case study of Window C showing offered input load, waiting requests, and ElasticKV's resident BASE ratio over time.}
	\label{fig_mooncake_a100_d20}
\end{figure}

Figure~\ref{fig_mooncake_a100_d20} repeats the production-trace experiment with $d=20$, reducing the offered arrival rate relative to the $d=16$ replay while preserving the same requests and relative arrival pattern. 

The same qualitative trends persist under the lighter replay. Exact-offload still incurs 211, 69, 212 native preemptions across A, B, C, respectively, while ElasticKV incurs none. Relative to Exact-offload, ElasticKV reduces mean TTFT by 29.1\%, 10.1\%, 23.4\% and mean waiting by 34.5\%, 15.1\%, 27.4\%, respectively. At this lighter load, completion throughput is already close to trace pacing, indicating that the remaining benefit primarily comes from reduced queueing and interruption rather than higher raw decoding speed. Observed BASE exposure is lower in the $d=20$ runs across all three windows, consistent with less sustained use of BASE under the lighter replay.

\subsection{Trace Replay on AMD MI355X GPUs}
\label{app_mi355x}

This section provides additional configuration details and results for the large-scale cross-platform evaluation in \S\ref{sec_amd}. We deploy Llama-3.1-70B-Instruct on eight AMD MI355X GPUs, each with 288\,GB HBM, and replay production traffic from the same Mooncake trace used in \S\ref{sec_real}.

\paragraph{Replay protocol and window selection.}
As in the single-A100 evaluation, we sample temporally separated regions from the early, middle, and late portions of the production trace. Here, we use three 600s windows rather than the 300s windows in App.~\ref{app_mooncake}. The large-scale replay uses a smaller dilation of $d=5$, therefore, a 300s window replays for only around 1500s. Using 600s windows increases the replay horizon to approximately 3000s and exposes the systems to a longer period of sustained production traffic while preserving the temporal structure within each selected region.

\begin{table}[t]
    \centering
    \caption{Mooncake production-trace windows used for the large-scale MI355X evaluation. The three 600s windows sample the early, middle, and late portions of the trace. }
    \label{tab_mi355x_windows}
    \small
    \setlength{\tabcolsep}{5pt}
    \begin{tabular}{cclrrr}
        \toprule
        Window & Region & Trace interval (s) &
        Requests & Input tokens & Output tokens \\
        \midrule
        A & Early  & [300,900) & 1810 & 24.34M & 0.627M \\
        B & Middle & [1500,2100) & 2093 & 23.45M & 0.731M \\
        C & Late   & [2700,3300) & 2207 & 25.48M & 0.720M \\
        \bottomrule
    \end{tabular}
\end{table}

Table~\ref{tab_mi355x_windows} reports the exact windows and workload volumes. Within each window, all systems receive the same requests in the same order and with the same arrival times and input/output lengths. Because the single-A100 and MI355X studies differ in model scale, hardware, window duration, and arrival-time dilation, we use the latter as a scale and cross-platform generalization test rather than as a direct cross-hardware speedup comparison.

\paragraph{Primary TP2$\times$DP4 deployment.}
Our primary configuration in \S\ref{sec_amd} uses four data-parallel (DP) model replicas, with each Llama-3.1-70B-Instruct replica tensor-parallelized (TP) across two GPUs. Table~\ref{tab_mi355x_tp2dp4} provides the exact per-window results underlying Fig.~\ref{fig_mi355x}. In addition to Eff@300, native request preemptions, and drain time shown in the figure, we also report mean TTFT as a conventional latency metric independent of the Eff@300 threshold, as well as BASE exposure to characterize the amount of mixed-fidelity execution used by ElasticKV. 

\begin{table}[t]
    \centering
    \caption{Detailed results for the TP2$\times$DP4 Mooncake replay on 8$\times$ AMD MI355X GPUs.}
    \label{tab_mi355x_tp2dp4}
    \small
    \setlength{\tabcolsep}{3.2pt}
    \begin{tabular}{clrrrrrr}
        \toprule
        Window & System &
        Eff@300 &
        Mean TTFT &
        P90 TTFT &
        Preempt. &
        Drain &
        BASE exp. (\%) \\
        \midrule

        \multirow{3}{*}{A}
        & Full-GPU      & 27.79\% & 272.49s & 924.63s & 649 & 1454s & -- \\
        & Exact-offload & 25.47\% &  84.42s & 251.64s & 450 &  665s & -- \\
        & ElasticKV     & 38.12\% &  40.16s & 118.63s &   0 &  299s & 4.84 \\
        \midrule

        \multirow{3}{*}{B}
        & Full-GPU      & 36.69\% & 96.79s & 356.03s & 484 & 477s & -- \\
        & Exact-offload & 35.88\% & 24.89s &  76.60s & 116 & 403s & -- \\
        & ElasticKV     & 55.90\% & 12.89s &  39.73s &   0 & 109s & 0.01 \\
        \midrule

        \multirow{3}{*}{C}
        & Full-GPU      & 37.92\% & 55.48s & 211.24s & 426 & 558s & -- \\
        & Exact-offload & 43.54\% & 35.79s &  96.82s & 183 & 515s & -- \\
        & ElasticKV     & 46.67\% & 18.01s &  55.72s &   0 &  59s & 0.02 \\
        \bottomrule
    \end{tabular}
\end{table}

The results are consistent across the three temporally separated windows. ElasticKV remains free of native request preemptions while improving the SLO-oriented effective-request ratio and substantially shortening the post-injection drain period. BASE exposure measures the fraction of attention-visible KV pairs executed in BASE and therefore differs from BASE residency or the number of fidelity transitions. The very low exposure in windows B and C indicates that substantial serving benefits can arise even when BASE is only rarely visible to attention execution. The mean-TTFT results further verify that the observed benefit is not specific to the Eff@300 threshold. Overall, this confirms that mixed-fidelity execution provides useful capacity elasticity when the model, accelerator platform, and distributed execution scale are all increased.

\begin{wrapfigure}{r}{.6\textwidth}
\vspace{-4mm}
\centerline{\includegraphics[width=\linewidth]{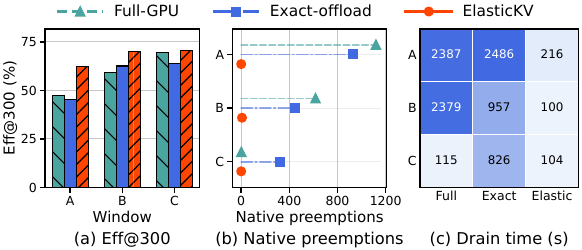}}
	\centering
	\caption{Production-trace replay on 8$\times$ AMD MI355X with Llama-3.1-70B-Instruct (TP1$\times$DP8).}
    \label{fig_app_mi355x}
\end{wrapfigure}
\paragraph{Alternative TP1$\times$DP8 topology.}
We further evaluate a substantially different parallel decomposition, TP1$\times$DP8, in which each GPU hosts one independent Llama-3.1-70B-Instruct replica and no intra-replica tensor parallelism is used. As shown in Fig.~\ref{fig_app_mi355x}, ElasticKV exhibits the same qualitative behavior under this topology. It nearly eliminates native request preemptions, with only seven events in total across the three-window replay. It maintains favorable Eff@300 and drain time relative to the baselines. We use this experiment as a topology-generalization check rather than a comparison of TP and DP efficiency. Together with the TP2$\times$DP4 deployment, it shows that ElasticKV's benefit is not specific to a single topology.

\subsection{Static FP8 as a Fixed-Fidelity Reference}
\label{app_fp8}

Static FP8 changes the target representation, but not the execution contract. Under a target-only runtime, FP8 becomes the single execution-ready TARGET. Its smaller footprint shifts the GPU-capacity boundary, but a request still cannot progress when the required FP8 TARGET is unavailable. In contrast, ElasticKV addresses a different dimension of the design space. It retains the native 16-bit representation as TARGET and introduces BASE as an additional executable state, allowing target fidelity to be recovered after progress has already resumed.

Table~\ref{tab_fp8} reports static FP8 as an end-to-end fixed-target reference under the same workload and GPU KV budget used in \S\ref{sec_concurrency}. We report the results at maximum concurrency of 38 and 36 on Llama-3.1-8B-Instruct and Qwen3-8B, respectively. On our NVIDIA A100 setup, default backend selection executes this configuration with FlashInfer~\citep{flashinfer}, while our FP16 and ElasticKV experiments use \texttt{TRITON\_ATTN} and \texttt{ELASTIC\_TRITON\_ATTN}, respectively. We therefore do not interpret the absolute latency or throughput differences as a backend-matched kernel comparison.

\begin{table}[t]
    \centering
    \small
    \caption{Serving performance comparison with static FP8. On our A100 setup, vLLM serves FP8 KV with FlashInfer, while ElasticKV uses our Triton-based elastic backend. The results are therefore not a backend-matched kernel comparison.}
    \label{tab_fp8}
    \setlength{\tabcolsep}{3.2pt}
    \begin{tabular}{llccrrr}
        \toprule
        Model &
        Setting &
        Executable KV states &
        Backend &
        P90 TTFT (s) &
        Out. tok/s &
        Preempt. \\
        \midrule

        \multirow{2}{*}{Llama-3.1-8B}
        & Static FP8
        & TARGET$_{\mathrm{FP8}}$
        & FlashInfer
        & 1.47
        & 1003.12
        & 0 \\
        & ElasticKV
        & TARGET$_{16}$ + BASE$_8$
        & Elastic Triton
        & 2.87
        & 914.25
        & 0 \\
        \midrule

        \multirow{2}{*}{Qwen3-8B}
        & Static FP8
        & TARGET$_{\mathrm{FP8}}$
        & FlashInfer
        & 1.50
        & 949.04
        & 0 \\
        & ElasticKV
        & TARGET$_{16}$ + BASE$_8$
        & Elastic Triton
        & 1.98
        & 854.92
        & 0 \\

        \bottomrule
    \end{tabular}
\end{table}

As expected, static FP8 provides a strong capacity-oriented operating point because its smaller target representation applies to the entire KV cache throughout execution. ElasticKV instead preserves the 16-bit target and invokes lower-fidelity execution only when needed to maintain progress. They represent different trade-offs rather than alternative implementations of the same mechanism.

More generally, the mixed-fidelity execution contract is orthogonal to the absolute precision of TARGET. A serving runtime with a lower-precision TARGET (e.g., an 8-bit representation) have the possibility to similarly introduce an additional lower-fidelity executable state (e.g., a 4-bit representation) using the proposed methodology of ElasticKV. Our current implementation instantiates this design with a 16-bit TARGET and an 8-bit BASE.

\subsection{Runtime Overhead Profiling}
\label{app_overhead}

\subsubsection{Profiling Setup and Methodology}

In \S\ref{sec_overhead}, we profile Llama-3.1-8B-Instruct on one NVIDIA A100 80GB GPU using a 20GB GPU KV budget, 600 requests, 4096-token inputs, 1024-token outputs, and maximum concurrency 38. This is the same controlled workload used in \S\ref{sec_concurrency}. At this operating point, ElasticKV remains preemption-free, while reaching an average BASE exposure of 14.7\%, providing a representative high-pressure setting in which mixed-fidelity execution and fidelity transitions are actively exercised.

\begin{wraptable}{r}{0.48\textwidth}
\vspace{-5mm}
    \centering
    \small
    \caption{Detailed breakdown of ElasticKV-specific critical-path work in the profiled representative high-pressure run. Cost is normalized by all steps.}
    \vspace{-1mm}
    \label{tab_overhead_detail}
    \setlength{\tabcolsep}{2pt}
    \begin{tabular}{lcc}
        \toprule
        Activity & Trigger (\%) & Cost (ms/step) \\
        \midrule
        \multicolumn{3}{l}{\textit{Online control}} \\
        \quad Shortage probe         & 78.67 & 0.072 \\
        \quad Demotion planning      & 11.06 & 0.175 \\
        \quad Promotion planning     & 100.00 & 0.023 \\
        \quad Restore-source prep.   & 100.00 & 0.021 \\
        \quad Async promotion issue  & 13.63 & 0.029 \\
        \midrule
        Sync. demotion               & 11.06 & 2.815 \\
        \midrule
        \multicolumn{3}{l}{\textit{Elastic metadata}} \\
        \quad Scheduler metadata     & 100.00 & 0.702 \\
        \quad Worker metadata        & 99.99  & 0.527 \\
        \quad Backend lowering       & 99.99  & 0.001 \\
        \midrule
        \textbf{Overall}             & --     & \textbf{4.365} \\
        \bottomrule
    \end{tabular}
    \vspace{-5mm}
\end{wraptable}
We measure only ElasticKV-specific work on the serving critical path and divide the overhead into three components.
\begin{itemize}[leftmargin=*,itemsep=2pt,topsep=3pt,parsep=0pt]
    \item \emph{Online control} includes shortage probing, demotion planning, reserve-aware promotion planning, restore-source preparation, and host-side asynchronous promotion issue.

    \item \emph{Synchronous demotion} measures the complete blocking path from invoking a TARGET-to-BASE transition until the physical blocks are committed and returned to the allocator-visible free pool.

    \item \emph{Elastic metadata} includes ElasticKV-specific scheduler metadata, worker-side mixed-fidelity metadata construction, and backend lowering, excluding native metadata required by vLLM.
\end{itemize}

For each activity, we report its invocation frequency and its accumulated latency normalized by the 16586 scheduling steps in the profiled run. Note that the CPU-to-GPU promotion transfer proceeds asynchronously and is excluded from the critical-path cost. Its host-side preparation and submission are already included in online control.

\subsubsection{Detailed Runtime Overhead Breakdown}


Table~\ref{tab_overhead_detail} shows that the online control logic is lightweight, adding only 0.32ms per scheduling step in total. Most of this cost comes from demotion planning, while shortage probing, promotion planning, restore-source preparation, and asynchronous promotion issue each contribute less than 0.1ms/step.

\begin{wraptable}{r}{0.33\textwidth}
    \centering
    \small
    \caption{Transition statistics for the profiled high-pressure run.}
    \label{tab_overhead_transition}
    \setlength{\tabcolsep}{2pt}
    \begin{tabular}{lr}
        \toprule
        Statistic & Value \\
        \midrule
        Scheduling steps          & 16586 \\
        BASE exposure             & 14.71\% \\
        Successful demotion batches & 1835 \\
        Demoted pairs             & 29427 \\
        Promotion issues          & 2260 \\
        Promoted pairs            & 25278 \\
        Native preemptions        & 0 \\
        \bottomrule
    \end{tabular}
    \vspace{-2mm}
\end{wraptable}
Synchronous demotion is the largest component. It is invoked on 11.06\% of scheduling steps and costs 25.44ms per invocation, but contributes only 2.815ms/step when amortized over all scheduling steps. Elastic metadata accounts for another 1.23ms/step, almost entirely from scheduler- and worker-side metadata construction. Backend lowering itself is negligible. Together, the three components account for 4.365ms of critical-path work per scheduling step.

Table~\ref{tab_overhead_transition} confirms that the profiled run substantially exercises both demotion and promotion while remaining preemption-free. All 1835 demotion batches completed successfully. The 29427 demoted pairs correspond to 16.04 pairs per demotion batch on average, closely matching the configured minimum batch size $B_{\min}^{d}=16$. This confirms that most triggered demotions are amortized at the configured minimum batch granularity rather than issued as tiny transitions.

\begin{wraptable}{r}{0.42\textwidth}
\vspace{-4mm}
    \centering
    \small
    \caption{Serving performance with runtime-overhead profiling disabled
    and enabled.}
    \label{tab_overhead_perturbation}
    \setlength{\tabcolsep}{2pt}
    \begin{tabular}{lrr}
        \toprule
        Metric & Profile off & Profile on \\
        \midrule
        Duration (s)              & 673.01 & 673.08 \\
        Mean TPOT (ms)            & 39.39  & 39.38 \\
        Output throughput (tok/s) & 912.92 & 912.82 \\
        Native preemptions        & 0      & 0 \\
        \bottomrule
    \end{tabular}
\end{wraptable}
We additionally verify that the profiling instrumentation itself does not measurably perturb serving by running the identical workload once with profiling disabled and once with profiling enabled. Table~\ref{tab_overhead_perturbation} shows nearly identical end-to-end behavior across the two runs. Enabling profiling changes benchmark duration by only 0.01\%, mean TPOT by $-0.03$\%, and output-token throughput by $-0.01$\%, indicating no serving perturbation in this paired check.

\subsection{Ablation Settings}
\label{app_ablation}

The ablation study in \S\ref{sec_overhead} uses the same high-pressure setting as the runtime overhead analysis described in App.~\ref{app_overhead}. All variants use the same workload, pair-structured KV layout, BASE representation, transition engine, and serving configuration. Each variant changes only the mechanism described below.
\begin{itemize}[leftmargin=*,itemsep=2pt,topsep=3pt,parsep=0pt]
    \item \emph{Watermark demotion} replaces ElasticKV's shortage-aware demotion trigger with a fixed free-capacity watermark. Demotion is triggered when available GPU KV capacity falls below the watermark. 

    \item \emph{Eager promotion} disables the demotion reserve used by ElasticKV's reserve-aware promotion policy, effectively setting the protected reserve to zero when computing the promotion budget. Promotion eligibility and transfer-slack checks remain unchanged.

    \item \emph{No promotion} disables BASE-to-TARGET promotion. Once a pair is demoted to BASE, it remains in BASE until its normal lifetime ends. 

    \item \emph{General elastic path} disables the structured mixed-fidelity attention path. Every mixed-fidelity batch is routed through the general pair-mode elastic implementation. TARGET-only batches continue to use the native Triton path.
\end{itemize}
\clearpage

\end{document}